\documentclass[conference,a4paper,final]{IEEEtran}
\IEEEoverridecommandlockouts

\usepackage[utf8]{inputenc}
\usepackage[T1]{fontenc}
\usepackage[english]{babel}
\usepackage{microtype}
\usepackage{blindtext}

\newcommand{\mycitestyle}{ieee}
\IfFileExists{ieee-comp.cbx}{
	\renewcommand{\mycitestyle}{ieee-comp}
}{}
\usepackage[
	style=ieee,
	citestyle=\mycitestyle,
	backend=biber
]{biblatex}
\usepackage[pdftex]{graphicx}
\graphicspath{{../../fig/}{../../build/figures}}

\usepackage{pgfplots}
\pgfplotsset{compat=1.16}

\usepackage[acronym]{glossaries}
\newglossaryentry{asic}
{
	name=ASIC,
	description={ASIC}
}

\newglossaryentry{bss2}
{
	name=\mbox{BrainScaleS-2},
	description={BSS-2 system}
}

\newglossaryentry{fpga}
{
	name=FPGA,
	description={FPGA}
}

\newglossaryentry{pcb}
{
	name=PCB,
	description={Printed Circuit Board}
}

\newacronym{lif}{LIF}{leaky-integrate-and-fire}
\newacronym{snn}{SNN}{spiking neural network}
\newacronym{ild}{ILD}{interaural level difference}
\newacronym{itd}{ITD}{interaural time difference}
\newacronym{ppu}{PPU}{plasticity processing unit}
 
\usepackage{booktabs}
\usepackage{nicefrac}
\usepackage[obeyFinal]{todonotes}
\usepackage{csquotes}

\usepackage{siunitx}
\usepackage{tikz}
\usepackage{circuitikz}
\usetikzlibrary{decorations.pathreplacing,decorations.text}
\usetikzlibrary{calc}
\usetikzlibrary{fit}

\tikzstyle{panel} = [
inner sep=0pt,
]
\tikzstyle{panellabel} = [
anchor=north west,
inner sep=0pt,
font={\bfseries\sffamily},
]

\usepackage{listings}
\usepackage{algorithm}
\usepackage{algpseudocodex}
\usepackage{enumitem}

\PassOptionsToPackage{hyphens}{url}
\usepackage[
	hidelinks,
]{hyperref}
\usepackage[capitalise,nameinlink]{cleveref}

\newcommand{\todoin}[1]{\todo[inline,color=yellow]{#1}}

\renewbibmacro{finentry}{\iffieldequalstr{entrykey}{mueller2022}{\finentry\newpage}
	{\finentry}}

\newcommand\submittedtext{\footnotesize
	\textcopyright\,2026 IEEE\@.
	Personal use of this material is permitted.
	Permission from IEEE must be obtained for all other uses, in any current or future media, including reprinting/republishing this material for advertising or promotional purposes, creating new collective works, for resale or redistribution to servers or lists, or reuse of any copyrighted component of this work in other works.
}

\newcommand\submittednotice{\begin{tikzpicture}[remember picture,overlay]
	\node[anchor=south,yshift=30pt] at (current page.south) {\fbox{\parbox{\dimexpr0.82\textwidth-\fboxsep-\fboxrule\relax}{\submittedtext}}};
	\end{tikzpicture}}

\begin{document}

\title{
	Real-time processing of analog signals on accelerated neuromorphic hardware
}

\author{\IEEEauthorblockN{Yannik Stradmann\IEEEauthorrefmark{3},
		Johannes Schemmel\IEEEauthorrefmark{3},
		Mihai A. Petrovici\IEEEauthorrefmark{2},
		and
		Laura Kriener\IEEEauthorrefmark{4}\IEEEauthorrefmark{2}
	}
	\vspace{2mm}

	\IEEEauthorblockA{\IEEEauthorrefmark{3}\,Institute of Computer Engineering, Heidelberg University, Heidelberg, Germany\\
		\IEEEauthorrefmark{2}\,Department of Physiology, University of Bern, Bern, Switzerland\\
		\IEEEauthorrefmark{4}\,Institute of Neuroinformatics, University of Zürich and ETH Zürich, Zürich, Switzerland
	}\vspace{-2mm}

	\thanks{
		The presented work has received funding from
		the European Union's Horizon 2020 research and innovation programme under grant agreements 785907 and 945539 (Human Brain Project) and Horizon Europe grant agreement 101147319 (EBRAINS 2.0),
		the \foreignlanguage{ngerman}{Deutsche Forschungsgemeinschaft} (DFG, German Research Foundation) under Germany's Excellence Strategy EXC 2181/1-390900948 (Heidelberg STRUCTURES Excellence Cluster),
		the Swiss National Science Foundation under the Starting Grant Project UNITE (TMSGI2-211461),
		and the VolkswagenStiftung under the CLAM Project (9C854).
		We would like to express our gratitude for the ongoing support from the Manfred Stärk Foundation.
		The presented project is the continuation of an initial implementation started during the CapoCaccia Workshop toward Neuromorphic Intelligence 2023.
	}}

\maketitle

\submittednotice

\begin{abstract}
	Sensory processing with neuromorphic systems is typically done by using either event-based sensors or translating input signals to spikes before presenting them to the neuromorphic processor.
Here, we offer an alternative approach:
direct analog signal injection eliminates superfluous and power-intensive analog-to-digital and digital-to-analog conversions, making it particularly suitable for efficient near-sensor processing.
We demonstrate this by using the accelerated \gls{bss2} mixed-signal neuromorphic research platform and interfacing it directly to microphones and a servo-motor-driven actuator.
Utilizing \gls{bss2}'s \num{1000}-fold acceleration factor, we employ a spiking neural network to transform interaural time differences into a spatial code and thereby predict the location of sound sources.
Our primary contributions are the first demonstrations of direct, continuous-valued sensor data injection into the analog compute units of the BrainScaleS-2 ASIC, and actuator control using its embedded microprocessors.
This enables a fully on-chip processing pipeline---from sensory input handling, via spiking neural network processing to physical action.
We showcase this by programming the system to localize and align a servo motor with the spatial direction of transient noise peaks in real-time.
 \end{abstract}


\section{Introduction}\label{sec:introduction}

Neuromorphic computing employs principles from neuroscience on both algorithmic and technological level to research efficient, next generation information processing.
Current conventional processors are usually limited by their energy budget, which---in case of von Neumann machines---is dominated by communication cost for most workloads~\parencite{horowitz2014computing}.
Traditionally, neuromorphic computing promises to overcome this bottleneck through event-based data exchange and in-memory computation, thereby reducing the data in volume and making it cheaper to access.

In this work, we explore an alternative, less general, option to achieve more efficient sensory processing.
Especially given the high computational demand for calculating neuronal and synaptic interactions, many neuromorphic systems follow their biological archetype and use analog processing elements at their core~\parencite{indiveri2011,benjamin2014neurogrid,moradi2018scalable,neckar2019,pehle2022}.
When interacting with the physical environment, this analog nature allows us to skip superfluous analog-to-digital and digital-to-analog conversions by interfacing suitable continuous-valued sensor signals directly to the analog compute units.
The increasingly lossy nature of analog communication over distance, however, makes this approach primarily suitable for near-sensor processing.

\begin{figure}[t!]
	\vspace{1mm}
	\centerline{
\includegraphics[trim={0mm 0mm 0mm 10mm},clip,width=\columnwidth]{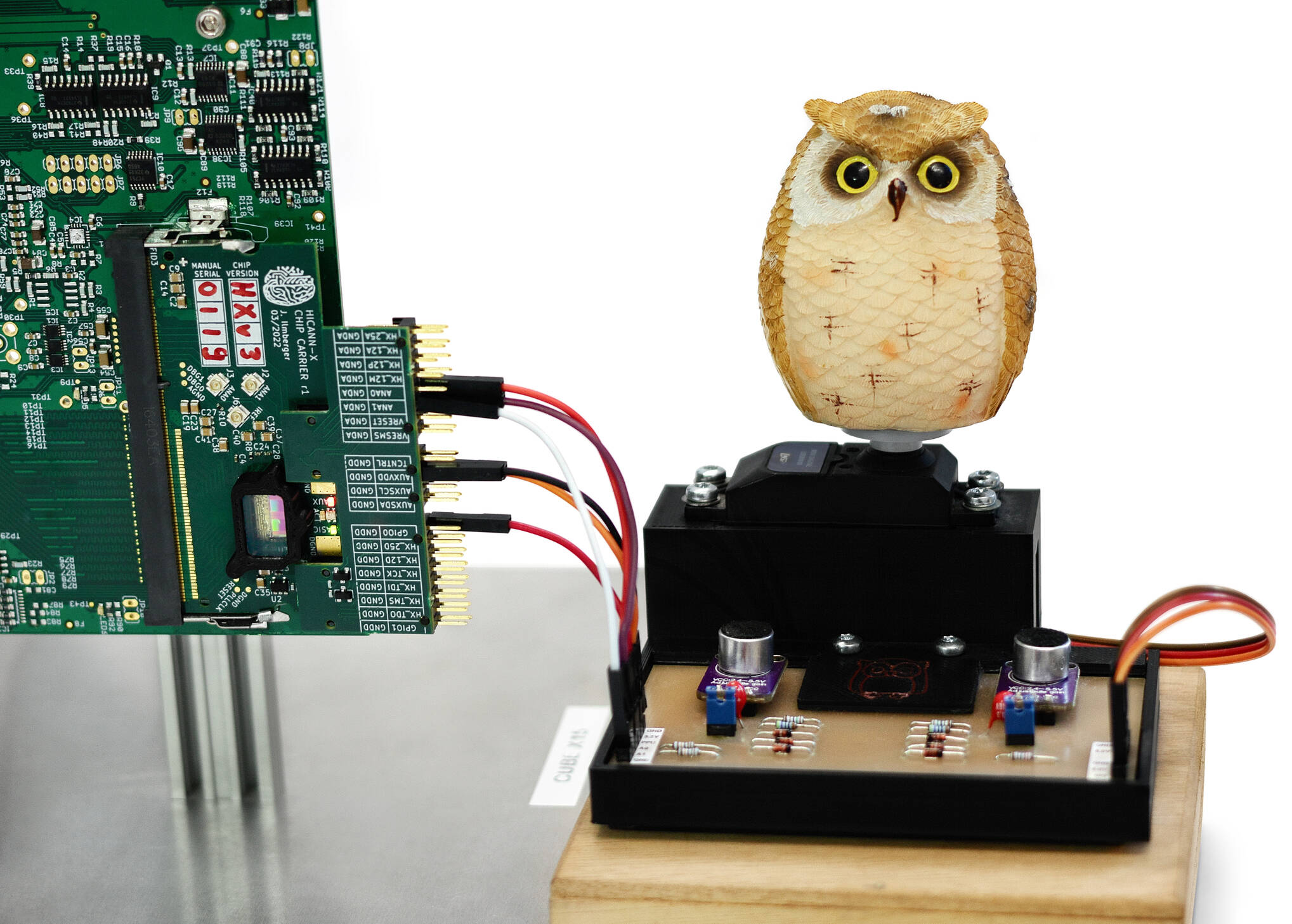}
	}
	\caption{
		Photograph of the presented demonstrator.
		The \gls{bss2} \gls{asic} is visible inside the black box on the left;
		it connects to a small \gls{pcb} with two microphones (silver capsules), and preprocessing circuits on the right.
		The owl figure turns, driven by a servo motor, toward the direction of detected sound sources.
	}
	\label{fig:system}
\end{figure}

Using the example of a simple sound localization task, we demonstrate the viability of this concept on the \gls{bss2} research platform~\parencite{pehle2022}, a mixed-signal neuromorphic system which allows direct analog access to any of its \num{512} individually configurable neuron circuits.
With an acceleration factor of \num{1000} compared to biological systems, \gls{bss2} features neuronal time scales in the microsecond range.
This inherently high temporal resolution of the system allows us to deploy a \gls{snn} that implements a strongly simplified version of the sound localization mechanism found in vertebrates, while still being able to localize sound sources from ear distances of \SI{10}{\centi\meter} and below.

The contributions of this work are twofold:
For the first time, we demonstrate the injection of continuous-valued sensor data directly into the analog computational elements of the \gls{bss2} \gls{asic}.
By additionally employing the chip's embedded microprocessors for output evaluation, this allows for a fully on-chip processing pipeline---from sensory input handling to actuator control.
We showcase these advances on the example of a simple sound localization task, where the system is programmed to align a servo motor with the spatial direction of transient noise peaks (\cref{fig:system}).
 \section{Methods}\label{sec:methods}

\begin{figure}
	\centerline{
		\includegraphics[width=0.8\columnwidth]{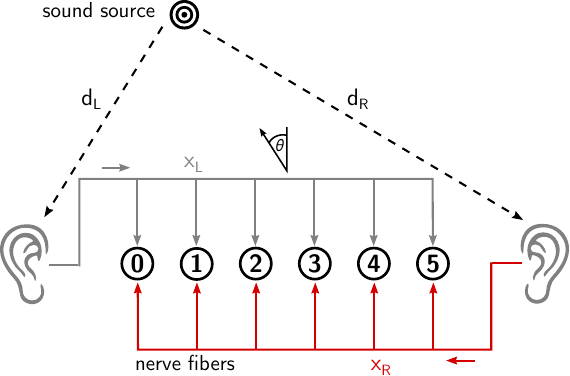}
	}
	\caption{
		Schematic illustration of the Jeffress model for sound localization.
		The difference in distance of both ears to a sound source ($d_\text{L}-d_\text{R}$) induces a difference in the sound's arrival time, the \acrfull{itd}.
		From the ears, the sound signal is transported to coincidence neurons (black) via nerve fibers.
		Depending on the distance traveled on the nerve fibers ($x_\text{L/R}$), delays of different lengths are introduced.
		Only where the respective accumulated delays compensate the \acrshort{itd}, the coincidence neuron becomes active.
		\vspace{-3mm}
	}
	\label{fig:jeffress}
\end{figure}

The two main cues employed for sound localization in vertebrates are the \gls{ild} and the \gls{itd}.
The former describes a delta in sound amplitude when arriving at each ear, typically either due to a longer travel distance or even an acoustic shadow created by the subject's head; the latter describes distance-induced timing differences.
Here, we focus on sound source localization based on \glspl{itd}:
Any point-like source has a specific distance to a listener's left ($d_\text{L}$) and right ($d_\text{R}$) ear (\cref{fig:jeffress}).
At constant speed of sound $c_\text{S}$, the difference between $d_\text{L}$ and $d_\text{R}$ results in a difference in the arrival times of the sound at the two ears:
\begin{align}
	\text{ITD} = \frac{d_\text{L} - d_\text{R}}{c_\text{S}}
\end{align}
The \gls{itd} can be used to reconstruct the spatial angle to the source $\theta$.
For homogeneous wave fronts (sound source at infinite distance), this dependency is commonly approximated using Woodworth's formula~\parencite{woodworth1938}
\begin{align}
	\text{ITD} = \frac{r_\text{H}}{c_\text{S}} \left( \theta + \sin{\theta} \right).
	\label{eq:woodworth}
\end{align}
Here, $r_\text{H}$ denotes the subject's head radius, i.e., half of its inter-ear distance.
Its value also determines the frequency range in which \gls{itd}-based sound localization is feasible.

For humans to achieve an auditory spatial resolution of $\le\SI{1}{\degree}$, time differences below $\text{ITD}_{\SI{1}{\degree}} = \SI{8.9}{\micro\second}$ must be resolved~\parencite{mills1958}.
Vertebrates with smaller heads even face proportionately higher demands.
Such high temporal precision is typically challenging to achieve for neuromorphic hardware that operates on neuronal time scales of milliseconds and does not feature specialized components to model the auditory pathway.
While accelerated systems present their own set of challenges for real-time signal processing, in a setting with high temporal precision requirements they are at an advantage, as their inherent dynamics can be several orders of magnitude faster than those in hardware operating on biological time scales.

\begin{figure}[b]
	\vspace{-2mm}
	\centerline{
		\includegraphics[width=0.8\columnwidth]{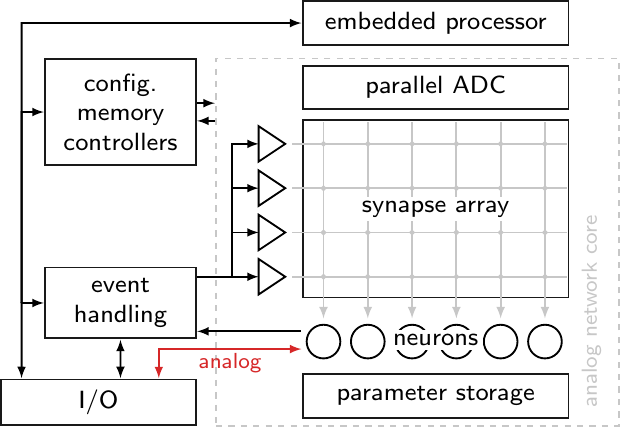}
	}
	\caption{
		Schematic overview over a single \gls{bss2} \gls{asic}, adapted from~\textcite{pehle2022}.
		The fully programmable synapse array allows the realization of near-arbitrary user-defined network topologies on the chip.
		The signal path used for injecting external, continuous-valued sensor data into the analog neuron circuits is highlighted in red.
	}
	\label{fig:bss2_schematic}
\end{figure}

\begin{figure*}
	\centerline{
		\includegraphics[width=0.9\textwidth]{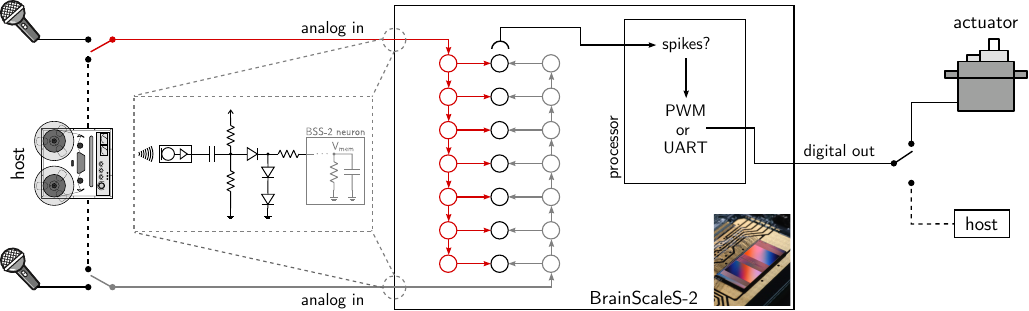}
	}
	\caption{
		Sketch of the full signal processing pipeline.
		Sound signals are either captured via two microphones or mimicked by the sound card of a host computer (left).
		The analog signals, after passive conditioning to protect the chip circuitry (inset), directly stimulate the \gls{bss2} neuron circuits.
		The on-chip network implements a simplified version of the Jeffress model where the nerve-fiber delays are implemented as chains of connected neurons.
		The spike counts of the coincidence neurons are read out by the embedded processor, which produces an output signal to either operate the actuator, or to be parsed by the host computer (right).
		\vspace{-3mm}
	}
	\label{fig:system_schematic}
\end{figure*}

This motivates our choice to use \gls{bss2}, a mixed-signal neuromorphic platform that has been developed to enable research in computational neuroscience, machine learning and computer science~\parencite{pehle2022,mueller2022}.
With an acceleration factor of \num{1000} compared to typical biological systems, it is particularly suited for long-running learning experiments and real-time interaction with fast physical systems~\parencite{stradmann2024}.
A single \gls{asic} features \num{512} analog neuron circuits emulating the Adaptive Exponential Leaky-Integrate-and-Fire neuron model~\parencite{brette2005}.
Each of these circuits receives input from \num{256} plastic synapses with \SI{6}{\bit} weight resolution.
Extensive analog and digital periphery enables fine-grained control over all model parameters, such as time constants, conductances, and potentials.
Two general-purpose embedded microprocessors per chip can act as system controllers or execute arbitrary, user-specified plasticity rules based on digital or analog observables---the latter being digitized by \num{1024} on-chip ADC channels.
\Cref{fig:bss2_schematic} depicts a simplified overview over the described components.

We have selected \gls{bss2} as the neuromorphic platform for this work due to three distinct features:
First, the system's acceleration factor allows us to construct neural networks that naturally operate on the fine time scales necessary for sound localization tasks.
Second, it provides direct, analog access to the neuron's membranes and therefore enables the injection of continuous-valued sensor signals without prior digitization or conversion to an event stream.
Third, the integrated embedded processors can generate actuator control signals and therefore enable a fully on-chip processing pipeline with three main stages (\cref{fig:system_schematic}):
\begin{enumerate}
	\item A stereo signal is captured by two microphones, pre-processed and injected into the on-chip input layer.
	\item The \gls{snn} processes this data and determines the \gls{itd} between the two signals.
	\item One of the embedded processors reads out the predicted \gls{itd} from the network's output layer and controls an actuator accordingly.
\end{enumerate}
In the following, we will describe these three steps in detail.

\subsection{Sensory input and analog preprocessing}\label{subsec:sensory-input}
The stereo audio signal used for sound localization is captured by two electret microphones, spaced \SI{51}{\milli\meter} apart.
Each channel runs through a corresponding preamplifier with adjustable gain to increase the signal amplitude and lower its impedance.

After the preamplifiers, the audio signal is centered around $\nicefrac{V_\text{DD}}{2}$ and therefore has a DC offset and amplitude of \SI{1.65}{\volt} each.
With the \gls{bss2} neurons operating in a voltage range from \SIrange{0}{1.2}{\volt}, the signals need minor passive preprocessing before they can be fed into the \gls{asic} (see inset in \cref{fig:system_schematic}):
As we can omit the negative half-wave for the task at hand, we high-pass the signal, set a new DC offset of \SI{0.8}{\volt} and filter any components below $\approx\SI{0.6}{\volt}$ through a diode connected in series.
The resulting signal resides between \SI{0.2}{\volt} and \SI{1.85}{\volt} and is subsequently clipped to $\approx\SI{1.2}{\volt}$ through two diodes in series to ground.
This limiter likewise ensures that no harmful signals can ever reach the delicate integrated circuits on the \gls{asic}.
Finally, we ensure a source impedance of at least $\SI{110}{\kilo\ohm}$.

These preprocessed audio signals are used to directly stimulate the membrane of two on-chip neurons.
Here, the \gls{bss2} software stack~\parencite{mueller2022} can automatically set the required analog routing between a neuron's membrane and one of two I/O pins of the \gls{asic} (\cref{listing:pynn_analogin,fig:bss2_schematic}).

For an automated performance evaluation of the full system, the microphone input can optionally be mocked by the line output of any common sound card:
By introducing an artificial shift between left and right channels of a stereo recording (\cref{fig:results}A), sounds arriving from different angles can be mimicked.
Since these mocked inputs are based on recordings with the very microphones of the physical system, they include any background noise and transducer non-linearities that occur in the interactive scenario.
By injecting them into the analog preprocessing board, we ensure the evaluation uses the identical signal conditioning, including potential signal degradation and noise characteristics of the physical setup.

\begin{lstlisting}[
	caption={The \gls{bss2} software stack allows to enable unbuffered, analog access to the neurons' membranes through a single PyNN~\parencite{davison2009pynn} \enquote{record} statement.
	Here, we select an input neuron of the network (first line) and ask the software to route its membrane directly to one of the \gls{asic}'s pins (second line).
	},
	captionpos=b,
	label={listing:pynn_analogin},
	basicstyle=\tt\footnotesize,
	numbers=none,
	language=Python,
	numberstyle=\footnotesize\color{black!30},
	stepnumber=1]
input_r = PopulationView(network, [0])
input_r.record("v", device="pad_0_unbuffered")
\end{lstlisting}

\subsection{Spiking neural network}\label{subsec:model}
The on-chip \gls{snn} for measuring the \gls{itd} of an incoming analog signal is inspired by Jeffress' model for sound localization in barn owls~\parencite{jeffress1948place} (\cref{fig:jeffress}):
The sound signals arriving at each ear are fed into chains of delay elements.
\todoin{think about: \enquote{, which creates a spatial representation of the input signal:}}
While propagating a distance $x_\text{L/R}$ along these delay chains, a stimulus is delayed by $\delta_\text{L/R}(x_\text{L/R})$.
Along this pair of chains, neurons are placed that receive input from both sides;
each of these neurons acts as a spatio-temporal coincidence detector and is sensitive to a specific $\text{ITD} = \delta_\text{R} - \delta_\text{L}$.

Biology achieves the chain's necessary microsecond-scale temporal resolution by employing nerve fibers with appropriate conductance as delay elements.
While reproducing these high-precision delay mechanisms in systems operating on biological time scales is challenging without specialized circuits, our choice of an accelerated hardware system significantly simplifies the task:
It allows us to model this propagation simply by using \gls{lif} neurons with exponential synaptic kernels, which naturally have time constants on the correct scale.
\todoin{MAP: Sketch PSP-induced delay and two-PSP-coincidence detection}
The implemented model therefore consists of two forward-projecting chains, where each neuron acts as a delay element, and an equally sized population of coincidence detection cells in between.

Once an analog stimulus of sufficient strength triggers the respective first neurons in each chain to fire, the signal propagates in a counter-directional manner (\cref{fig:system_schematic}).
Only at the coincidence detection neuron where the difference in accumulated delays along the chains counters the \gls{itd}, both chains are active at the same point in time and should thereby create sufficiently strong stimulation to cause an action potential.
Again, the accelerated nature of the neuron circuits makes this easy in our setup, because the neurons operate on time scales of microseconds.
Therefore, input timing differences in the microsecond range have a big impact on the neuron's output.
For neurons operating in real-time, however, more sophisticated processes are required to be sensitive to such small time differences (see for example~\cite[Chapter 12.5]{gerstner2002spiking}).
The activity of each coincidence detection unit therefore marks sound sources at a specific azimuthal angle in front of the stereo microphone.

\begin{algorithm}[b]
	\begin{algorithmic}[1]
		\While{\textbf{true}}
			\State accumulator $\gets 0$
			\State active\_neurons $\gets 0$
			\For{neuron\_id $\in \{\text{coincidence\_detectors\}}$}
				\If{\Call{GetEventCounter}{neuron\_id}}
					\State accumulator $\gets$ accumulator $+$ neuron\_id
					\State active\_neurons $\gets$ active\_neurons $+$ 1
				\EndIf
			\EndFor

			\If{\textbf{not} active\_neurons}
				\State \textbf{continue}
			\EndIf
			\State detected\_direction $\gets \text{accumulator} / \text{active\_neurons}$
			\State \Call{UpdateOutput}{detected\_direction}

			\State \Call{Sleep}{\SI{200}{\milli\second}}
			\State \Call{ResetEventCounters}{\,\!}
		\EndWhile
	\end{algorithmic}
	\caption{Coincidence detector readout program running on the embedded processor.}
	\label{alg:ppu_program}
\end{algorithm}

\subsection{Direction readout and actuator signal}\label{subsec:ppu_output}
To fulfill the goal of a fully on-chip processing pipeline, we employ one of the two embedded microprocessors on each \gls{bss2} \gls{asic} to generate actuator signals based on the \gls{snn}'s response.
While this processor has no access to the exact spike times of the neurons on the chip, it can read out the output spike counters of each of the coincidence neurons one after the other.
Therefore, in order to determine whether the sound localization network has detected an incoming sound, the microprocessor continuously polls these counters for new events (see \cref{alg:ppu_program}).
As each iteration of this program has a finite runtime (\SI{55}{\micro\second}), it is possible that within one iteration multiple coincidence neurons have fired.
In those cases, we average the IDs of the neurons that have produced an event and use this average value as the detected direction.
After a sound source has been recognized, the system updates the output value for the actuator, enters a dead time period of \SI{200}{\milli\second} to ensure state decay in the full pipeline and prepares for subsequent events by resetting all spike counters.
This detection scheme prefers early, high-confidence events by only recognizing those neurons that have spiked within the first loop iteration after a sound occurrence.
It also creates a bias based on the neurons' enumeration order, which we have, however, not found to be recognizable in practice.

For the system's output, two options are available:
Using a digital I/O pin, the embedded processor can either produce a direction-modulated PWM signal or use a software-defined UART implementation to output serial data.
The PWM signal can directly control off-the-shelf servo motors and is intended to be used for tracking a noise source in real-time.
Especially in combination with the pre-recorded audio input described in \cref{subsec:sensory-input}, the serial output is useful when performing fully automated sweeps for system performance evaluations.
 \begin{figure*}
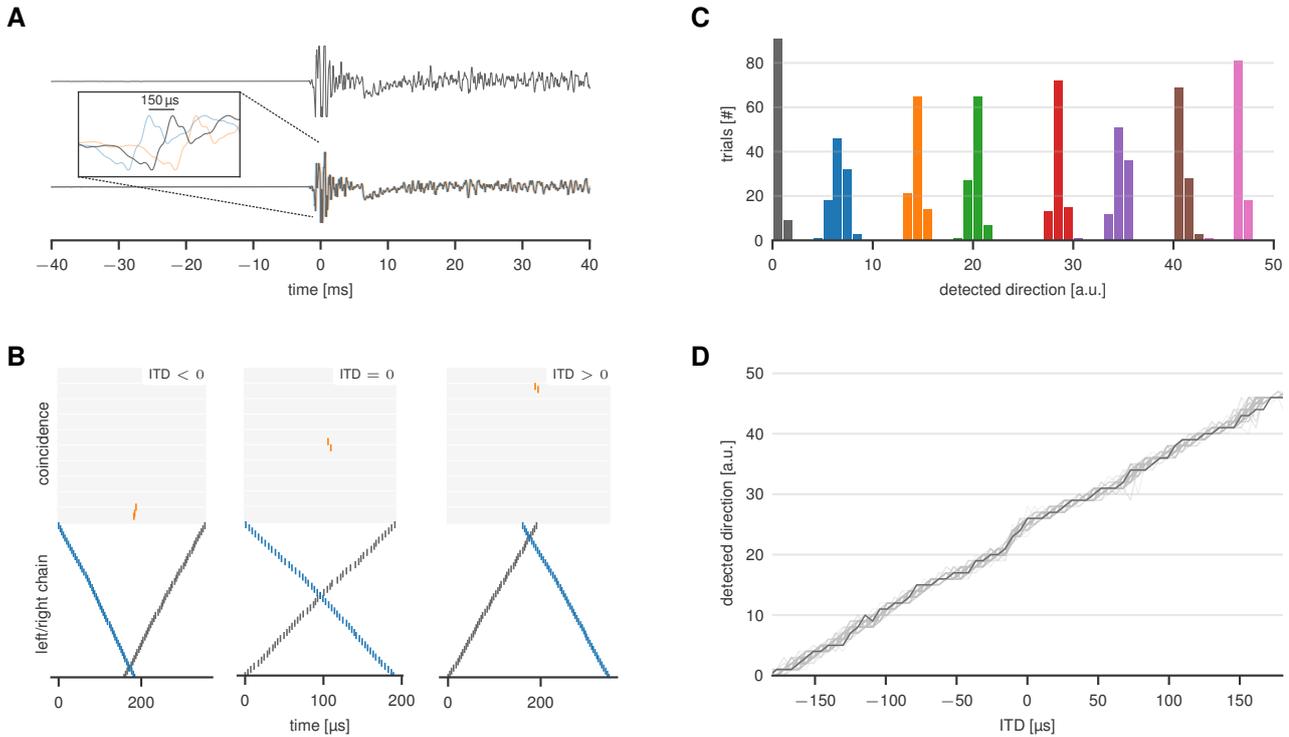

	\begin{tikzpicture}

		\node[panel, anchor=north west] (a) at (0, 0) {
			\input{clap_wave.pgf}
		};

		\node[panel, anchor=north west] (b) at (0, -4.5) {
			\input{spiketrain_left-first.pgf}
			\input{spiketrain_center.pgf}
			\input{spiketrain_right-first.pgf}
		};

		\node[panel, anchor=north west] (c) at (9, 0) {
			\input{histograms.pgf}
		};

		\node[panel, anchor=north west] (d) at (9, -4.5) {
			\input{delay_sweep.pgf}
		};

		\node[panellabel, anchor=east] at (a.north west) {A\vphantom{bp}};
		\node[panellabel, anchor=east] at (b.north west) {B\vphantom{bp}};
		\node[panellabel, anchor=east] at (c.north west) {C\vphantom{bp}};
		\node[panellabel, anchor=east] at (d.north west) {D\vphantom{bp}};
	\end{tikzpicture}
	\vspace{-2.5mm}

	\caption{(A) Stereo recording of a person clapping, using the microphones depicted in~\cref{fig:system}.
		One channel can be time delayed (orange) or advanced (blue) to mimic different sound source positions.
		(B) Rasterplots of the left (gray) and right (blue) delay chains (bottom) and coincidence detection neurons (top) for three different time delays between the left and right channel.
		(C) Variation over the sound direction as detected by the system for eight different \acrfullpl{itd}, each over \num{100} trials.
		(D) Detected sound direction over \acrshort{itd} between the input channels.
		For technical reasons, panel (B) has been recorded with artificially injected spikes to the respective first neuron.
		Panels (C) and (D) were recorded by playing back an audio recording of recorded claps (cf.~\cref{subsec:sensory-input}).
		\vspace{-3mm}
	}
	\label{fig:results}

\end{figure*}
 
\section{Results}\label{sec:results}

The described system can reliably predict the position of a sound with distinct onset, such as clapping or a bouncing ping-pong ball, in a room:
once a sufficiently strong transient sound is detected, it aligns the mounted figure to the noise source using an off-the-shelf servo motor.
With a measured latency of \SI{0.5}{\milli\second} (first signal transient to digital output), the processing time is virtually immediate and the motor's velocity dominates the time to reposition.

\begin{table}[h!]
	\vspace{-1mm}
	\centering
	\caption{Network parameters and performance}
	\label{tab:results}
	\begin{tabular}{lc}
		\toprule
		\textbf{Parameter} & {\textbf{Value}} \\
		\midrule
		delay stages & 50 \\
		synaptic time constant & \SI{15}{\micro\second} \\
		membrane time constant & \SI{15}{\micro\second} \\
		refractory time & \SI{0.5}{\milli\second} \\
		time delay per stage & \SI[separate-uncertainty=false]{3.8(8)}{\micro\second}\\
		time to output & \SI{0.5}{\milli\second}\\
		\bottomrule
	\end{tabular}
\end{table}
 
In \cref{fig:results}B, we depict the on-chip network dynamics of the two delay chains and the coincidence neurons for three stimuli: sound source located on the right, center, and left.
\todoin{MAP: mark in plots}
For either experiment, the coincidence neurons corresponding to the respective intersection of both delay chains become active.
With the exemplary \gls{lif} neuron parametrization listed in~\cref{tab:results}, we measure a mean time per delay stage of \SI{3.8(8)}{\micro\second}, indicating a theoretical spatial resolution of \SI{\le 1.5}{\degree} (\cref{eq:woodworth}).
Here, the target resolution and microphone distance constrain possible combinations of time constant settings and chain lengths.
In a perfectly tuned system, a single orange-marked coincidence neuron should spike for each of the three experiments.
As can be seen from the rasterplot, this is not the case for the current parametrization:
the temporal receptive field of the coincidence neurons is broader than ideal, reducing the resolution of the prediction (cf.~\cref{sec:discussion}).

For an automated evaluation of system performance over multiple \glspl{itd}, we bypass the microphones and play back a recorded clap with an adjustable, software-set inter-channel delay.
The corresponding stereo recording has been acquired from the mounted microphones and is displayed in \cref{fig:results}A.
Here, the bottom trace is shifted by a positive or negative time to encode signals from different spatial angles.
With a microphone distance of \SI{51}{\milli\meter}, a full half-space is encoded within \SI{\pm149}{\micro\second} time delay---a negligible shift in contrast to the sonic signature of even a short \enquote{clap}.

\Cref{fig:results}D depicts the overall system response when stimulated with such stereo recordings through an off-the-shelf sound card with \SI{192}{\kilo\hertz} sampling frequency.
The detected direction is measured by parsing the output of the embedded processor (\cref{alg:ppu_program}) and corresponds to the position of neuronal activity in the coincidence detection neurons.
Measured over \num{100} trials, all curves match the desired linear behavior, outliers rarely occur and are constrained locally.
For \num{8} \glspl{itd} of this measurement, \cref{fig:results}C shows the distribution of detected directions across all trials.
The distributions' widths correspond to the spread of the curves in \cref{fig:results}D, which we therefore assess to be \numrange{2}{4} units over the full receptive field.
 \section{Discussion}\label{sec:discussion}

In this work, we have employed the unique feature set of \gls{bss2} to explore real-time analog sensory processing and actuator control in a fully on-chip processing pipeline.
This has been facilitated by skipping the more commonly used spiking input interface and instead utilizing a direct analog input onto the \gls{bss2} compute units, employing the on-chip embedded microprocessor to generate actuator control signals, and by leveraging the inherent acceleration factor of \gls{bss2} for high temporal precision.
We have showcased this concept in a binaural sound localization task, where the implemented \gls{snn} represents a simplified version of biological models.

While similar sound localization tasks have already been demonstrated on various other neuromorphic systems, many of these previous solutions are either based on purpose-built \glspl{asic}~\parencite{lazzaro1989,park2013fast}, or rely on external---often \gls{fpga}-based---preprocessing to manage the strict timing requirements for resolving \glspl{itd}~\parencite{schoepe2023,kugler2007complete,dalmas2024review}.
Our approach, in contrast, employs \gls{bss2}'s acceleration factor to process the input signals natively within the \gls{snn}.
In future work, this will permit the exploration of more complex network topologies and enable the application of biological principles to previously unreachable time scales.
For example, backward inhibition in the delay chains would curb their response without relying on a prolonged refractory period, while winner-take-all circuits with lateral inhibition could be employed as coincidence detectors to sharpen the system's angular resolution.
Having implemented all delays as chains of freely configurable neurons allows us to further reduce the microphone distance, thereby extending the processing range into the ultrasonic domain.

Altogether, we establish \gls{bss2} as a platform for end-to-end, on-chip sensory processing---with inputs handled, computed, and evaluated entirely on the chip.
By enabling fast and flexible near-sensor analog computation, it unlocks biologically inspired solutions for a new class of problems with high demands on temporal precision.
 \printbibliography

\end{document}